\documentclass[conference]{IEEEtran}
\IEEEoverridecommandlockouts
\usepackage{amsmath,amssymb,amsfonts}
\usepackage{algorithmic}
\usepackage{graphicx}
\usepackage{textcomp}
\usepackage{xcolor}
\usepackage{subfigure}
\usepackage{balance} 
\def\BibTeX{{\rm B\kern-.05em{\sc i\kern-.025em b}\kern-.08em
    T\kern-.1667em\lower.7ex\hbox{E}\kern-.125emX}}
\usepackage{lipsum}

\title{Style Transfer Applied to Face Liveness Detection with User-Centered Models
¨
}

\makeatletter
\newcommand{\linebreakand}{%
  \end{@IEEEauthorhalign}
  \hfill\mbox{}\par
  \mbox{}\hfill\begin{@IEEEauthorhalign}
}
\makeatother

\author{
    \IEEEauthorblockN{Israel A. Laurensi R.}
    \IEEEauthorblockA{\textit{Post-Graduate Program in Computer Science (PPGIa)} \\
    \textit{Pontifical Catholic University of Parana (PUCPR)}\\
    Curitiba, PR, Brazil \\
    israel.rosa@pucpr.edu.br}
    \and
    \IEEEauthorblockN{Luciana T. Menon}
    \IEEEauthorblockA{\textit{Post-Graduate Program in Computer Science (PPGIa)} \\
    \textit{Pontifical Catholic University of Parana (PUCPR)}\\
    Curitiba, PR, Brazil \\
    luciana.menon@pucpr.edu.br}
    \linebreakand 
    \IEEEauthorblockN{Manoel Camillo O. Penna N.}
    \IEEEauthorblockA{\textit{Post-Graduate Program in Computer Science (PPGIa)} \\
    \textit{Pontifical Catholic University of Parana (PUCPR)}\\
    Curitiba, PR, Brazil \\
    penna@ppgia.pucpr.br}
    \and
    \IEEEauthorblockN{Alessandro L. Koerich}
    \IEEEauthorblockA{\textit{Department of Software and IT Engineering} \\
    \textit{\'{E}cole de Technologie Sup\'{e}rieure (\'{E}TS)}\\
    Montr\'{e}al, Canada \\
    alessandro.koerich@etsmtl.ca}
    \linebreakand
    \IEEEauthorblockN{Alceu S. Britto Jr.}
    \IEEEauthorblockA{\textit{Post-Graduate Program in Computer Science (PPGIa)} \\
    \textit{Pontifical Catholic University of Parana (PUCPR)}\\
    Curitiba, PR, Brazil \\
    alceu@ppgia.pucpr.br}
}

\begin{document}

\maketitle

\begin{abstract}

This paper proposes a face anti-spoofing user-centered model (FAS-UCM). The major difficulty, in this case, is obtaining fraudulent images from all users to train the models. To overcome this problem, the proposed method is divided in three main parts: generation of new spoof images, based on style transfer and spoof image representation models; training of a Convolutional Neural Network (CNN) for liveness detection; evaluation of the live and spoof testing images for each subject. The generalization of the CNN to perform style transfer has shown promising qualitative results. Preliminary results have shown that the proposed method is capable of distinguishing between live and spoof images on the SiW database, with an average classification error rate of 0.22.

\end{abstract}


\section{Introduction}
With the growth of technology and the advance of computer vision techniques and machine learning, facial biometry has been receiving special attention in the last few years \cite{facenet_schroff2015, deepface_Taigman2014, terceiro_2969033.2969049}. Not far from that, the ease of implementation and integration of facial biometric systems brings the concern with the security of these solutions. More specifically, when we consider facial verification, a major concern arises in regard to authenticity, in which one person tries to obtain access as another person. The problem addressed in this paper is the liveness detection on a face image, which means determining if there is really a living person in front of a camera and not an attempt to identity fraud by presenting a photo or a video in order to obtain improper access. Therefore, it is expected that a face anti-spoofing (FAS) system should be able to distinguish an image that does not have a fraud attempt from one that does have it. Thus, a given solution must be able to receive images captured from various sources (smartphones, webcams, professional cameras, etc.) and perform the classification of these images as authentic or fraudulent, which are defined from now on as spoof images.

Using machine learning algorithms to tackle the problem of face liveness detection requires examples of authentic images as well as fraudulent images. Several benchmark databases have been released in the last years in the context of face liveness detection \cite{NUAA_2010, casia_Zhang2012, replayattack_Chingovska_BIOSIG-2012, siw_DBLP:journals/corr/abs-1803-11097}, with the objective of providing training and test data to solve the problem linked to authenticity in facial recognition.

Although there are many spoof databases, they are not always representative enough for a real application. Several face liveness detection methods have been proposed, however, their results hardly beat random classifiers \cite{survey_Souza:2018:FDW:3223062.3223240}. In addition, it is observed that robust classifiers such as deep neural networks, often learn not only the spoof representation, but also facial characteristics of the subject present on the database. An interesting idea is to circumvent this problem is to create user-centered liveness detection models. However, the major difficulty in this case is obtaining spoof images from all subjects. In a real-world scenario, it is impracticable to ask a subject to provide examples of fraudulent images of himself. Therefore, it is important to create a method for generating these fraudulent images automatically.

In this paper we propose an approach for generating fraudulent face images from authentic ones based on the idea of style transferring, and use both authentic and fraudulent images to build user-centered face liveness detection models based on convolution neural networks (CNNs). For such an aim, we use the CNN-based approach proposed by Gatys et al. \cite{cnnopt1_gatys2015neural} that creates artistic images of high perceptual quality. Even if their main purpose is to create artistic images, we adapt their approach to create more secure facial biometric systems. Therefore, we use the style transfer techniques to create dynamically fraudulent images from real subjects. In addition, the idea of making the liveness detection user-centered brings new results in the context of facial biometrics and liveness detection.

The remainder of the paper is organized as follows: In Section II, the theoretical background of style transfer and data augmentation on facial biometrics is given. The proposed method is presented in Section III. Results and concluding remarks are given in Sections IV and V, respectively.
\section{Related Work}
Generative Adversarial Networks (GANs) and Convolutional Neural Networks (CNNs) have been extensively used in the process of generating new images given a specific database as a reference. The modeling of new images can be learned from the probability distribution of any set of images \cite{gan1_NIPS2014_5423}. This process can be perceived in the literature in applications such as the generation of new images \cite{gan3_karras2019style}, the transfer of styles from one set of images to another \cite{gan8_zhu2017unpaired}, the modeling of new images combining features in the discriminative space \cite{gan2_radford2015unsupervised}, among others.

In the context of anti-fraud in facial biometric systems, there are few studies that report the use of GANs as a countermeasure to presentation attacks. However, some studies report methods for face generation based on certain given attributes \cite{gan4_wang2018attributes}, methods based on data augmentation techniques \cite{gan5_choe2017face}, and methods based on the use of more than one GAN to generate facial images and analyze specific attributes in respect to qualitative measures \cite{gan6_zhao20183d}. Some authors have used GANs as an auxiliary step in the decision process to distinguish facial images from real and spoof. Jourabloo et al. \cite{gan7_Jourabloo_2018_ECCV} presented a GAN model to perform image denoising, with the purpose of mapping the noise present in spoof images and recreating a real image from a spoof image. According to Jourabloo et al. \cite{gan7_Jourabloo_2018_ECCV}, it is necessary that the neural networks used to create the GAN be able to capture different types of attacks -- which produce different types of noise in the images -- and, from the noise subtraction present in the image, an estimate of the real image can be generated, in the light that actual images would not produce such spoof noise. The experiments were conducted on the Oulu-NPU \cite{OULU_NPU_2017}, CASIA \cite{casia_Zhang2012} and Replay-Attack \cite{replayattack_Chingovska_BIOSIG-2012} databases. They achieved a half-total error rate (HTER) of 28.5\%, in a cross-database evaluation using the CASIA for training and Replay-Attack for test, and a HTER of 41.1\% when using Replay-Attack for training and CASIA for test.

Rehman et al. \cite{cnn1_Rehman2019} used a CNN-based auto-encoder (DNG) for face generation. The CNN was used to aid in the classification of real and spoof images, using the fusion of weights in the convolutional layers learned in the process of auto-encoding. The discrepancy between the histograms of the real images and the images generated by the DNG shows that the learning of the DNG network can be used to aid in the classification of real and spoof images. The cross-database results were 11.26\% HTER using CASIA \cite{casia_Zhang2012} for training and Replay-Attack \cite{replayattack_Chingovska_BIOSIG-2012} for test.

Liu et al. \cite{siw_DBLP:journals/corr/abs-1803-11097} proposed a CNN-RNN model to estimate the face depth maps and rPPG signals to distinguish live and spoof faces. Depth maps are a representation of the 3D shape of the face in a 2D image, which shows the depth information of different facial area. The rPPG signal is related to the intensity changes of facial skin over time, which are highly correlated with the blood flow. The authors proposed a method consisting of two deep networks, a CNN part that estimates the feature map of each frame and a recurrent neural network (RNN) that evaluates the temporal variability across the feature maps of a sequence of frames. Liu et al. \cite{siw_DBLP:journals/corr/abs-1803-11097} achieved an average classification error rate (ACER) of 3.58\% in the SiW database.

Wang et al. \cite{r46_DBLP:journals/corr/abs-1811-05118} combined temporal motion and facial depth to discriminate between living and spoofing faces. Their proposed model consists of two modules: (i) the single-frame part, which estimates the depth map from an individual frame; (ii) the multi-frame part which consists of a depth supervised neural network architecture with optical flow guided feature block (OFFB) and convolution gated recurrent units (ConvGRU) to combine short-term and long-term motion extractors. The authors achieved a 0.73\% ACER on the intra-database protocol in the SiW database.

Zhang et al. \cite{r40_DBLP:journals/corr/abs-1812-00408} proposed a multi-stream method based on ResNet-18 classification network. Their method consists of three blocks, each block uses ResNet-18 as backbone and extract features of each dataset modalities. RGB, Depth and IR data are learned separately, one in each stream. Then, these features from different modalities are fused via the squeeze and excitation fusion module to learn more discriminative features and perform cooperated decisions. Their protocol consists of training the model on their proposed dataset CASIA-SURF, fine-tune it on the target training database, and test on the target testing set. Zhang et al. \cite{r40_DBLP:journals/corr/abs-1812-00408} reported a 0.80\% ACER on the SiW database, following the intra-database protocol.

Zhao et al. \cite{rOurs_DBLP:journals/corr/abs-1904-12490} proposed a meta-learning method to adapt a neural network to distinguish between live and spoof faces with few images. Their method proposes two neural network approaches: (i) a neural network for live and spoof classification supervision; (ii) a neural network to perform depth map regression. Zhao et al. \cite{rOurs_DBLP:journals/corr/abs-1904-12490} performed cross-database tests on Oulu-NPU, MSU-MFSD, SiW and Replay-Attack. The reported intra-database tests resulted in an 0.51\% ACER in the SiW database.

Concerning the transfer of style from one set of images to another, Zhu et al. \cite{gan8_zhu2017unpaired} demonstrate the use of a GAN for the learning and mapping of the distribution of a set of images, in order to apply the learned distribution to new images. The problem addressed by the authors concerns the context of translation from one image to another -- which aims to capture a mapping of an input image and an output image (pairs) -- considering, however, the absence of the output images, given the GAN the task of learning a generic mapping (within specific contexts) given the input images \cite{gan8_zhu2017unpaired}.

Gatys et al. \cite{cnnopt1_gatys2015neural} presented a neural algorithm for style transfer based on the extraction of image style through convolutional layers. The authors showed that the deeper the convolutional layers, the more the content of the image and the artistic style could be separated -- and, as a result, more the artistic style could be extracted from the input image. Similar to this, the higher layers of the CNN can generate more robust, sharp and detailed artistic styles images \cite{cnnopt1_gatys2015neural}. Johnson et al. \cite{cnnopt2_johnson2016perceptual} brought to light optimization to the neural algorithm proposed by Gatys et al. \cite{cnnopt1_gatys2015neural}, where the neural feed-forward network was trained with perceptual loss, instead of a per-pixel loss. Such an optimization had similar qualitative results in regards to the artistic style transfer, with three orders of magnitude faster \cite{cnnopt2_johnson2016perceptual}. The optimization proposed by Ulyanov et al. \cite{cnnopt3_ulyanov2016instance} also showed that instance normalization could be applied to the CNN with improved results over batch normalization, in training and testing time.

\section{User-Centered Models}
\subsection{Database}

The database used in this paper was the Spoof in the Wild (SiW) database \cite{siw_DBLP:journals/corr/abs-1803-11097}. It was introduced in 2018 and consists of 165 subjects in 4478 different videos with 1080P HD resolution. Different sections were recorded to capture the videos, varying the participants as well as the lighting. The following types of presentation attacks are present in the database: photos of printed photos; presentation attack using cell phones; presentation attacks using monitor screens; and presentation attacks using tablet screens. The live videos were captured with two high-quality cameras (Canon EOS T6 and Logitech C920 webcam), in four different sessions: (i) subjects were asked to move their head with varying distances to the camera; (ii) subjects move yaw angle of the head varying from 90$^\circ$ to -90$^\circ$, with different facial expressions; (iii) and (iv) same movements as in (i) and (ii) but with variance in the light source illuminating the subject's face and changing orientation during the video capture, respectively.

\begin{figure}[htpb]
\centering
    \subfigure[]{\includegraphics[width=0.11\textwidth]{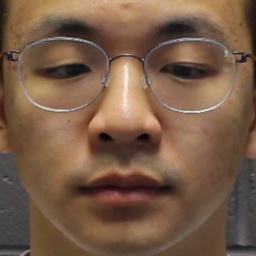}}
    \subfigure[]{\includegraphics[width=0.11\textwidth]{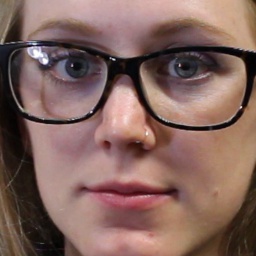}}
    \subfigure[]{\includegraphics[width=0.11\textwidth]{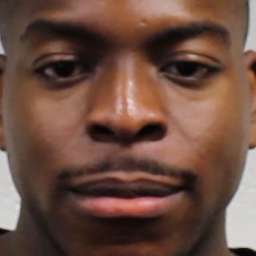}}
    \subfigure[]{\includegraphics[width=0.11\textwidth]{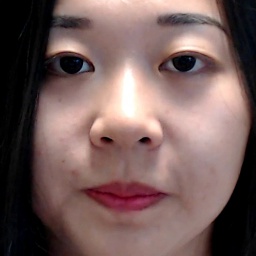}}\break
    \subfigure[]{\includegraphics[width=0.11\textwidth]{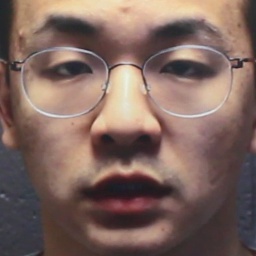}}
    \subfigure[]{\includegraphics[width=0.11\textwidth]{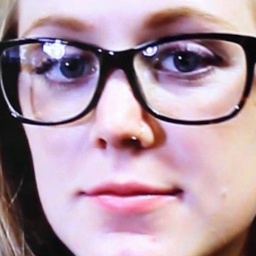}}
    \subfigure[]{\includegraphics[width=0.11\textwidth]{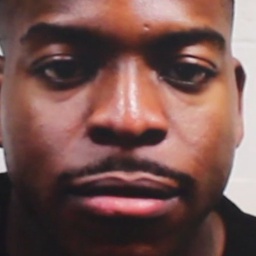}}
    \subfigure[]{\includegraphics[width=0.11\textwidth]{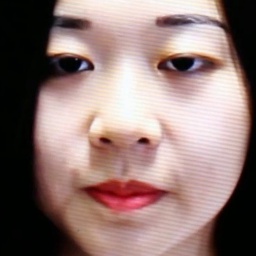}}

    \caption[]{Face samples from the SiW database: (a)--(d) live samples; (e)--(h) spoof samples.}
    \label{fig:exemplos-SIW}
\end{figure}

For facial extraction, we used the Dlib library \cite{dlib09}, with height and width for facial images of 256$\times$256 pixels and a margin of 0.1. Fig \ref{fig:exemplos-SIW} shows some samples of the extracted faces from the SiW database. The same face extraction protocol was applied to both live and spoof images. All frames from the videos were used to extract faces. In total, around 1.1M live image faces were obtained and around 1.4M spoof image faces. A higher number of spoof images is due to the different presentation attacks present in the database, given that, for each live video a number of different spoof videos were generated in the SiW database.

\subsection{Proposed Method}
The proposed face anti-spoofing method is divided in two main parts: (i) generation of new spoof images based on style transfer of spoof images representation models; (ii) training of a CNN model for live and spoof image classification.

\subsubsection{Style Transfer} 
\begin{figure}[htpb]
\centering
    \subfigure[]{\includegraphics[width=0.08\textwidth]{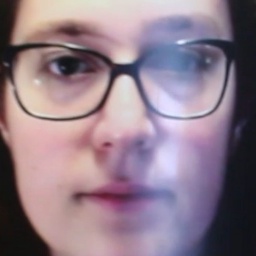}}
    \subfigure[]{\includegraphics[width=0.08\textwidth]{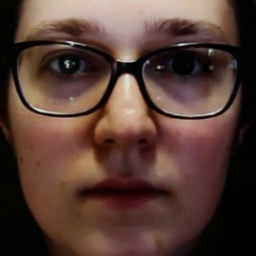}}
    \subfigure[]{\includegraphics[width=0.08\textwidth]{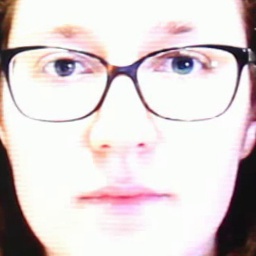}}
    \subfigure[]{\includegraphics[width=0.08\textwidth]{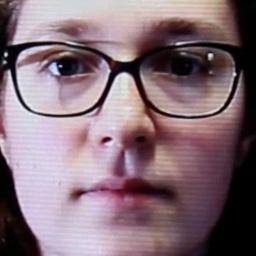}}
    \subfigure[]{\includegraphics[width=0.08\textwidth]{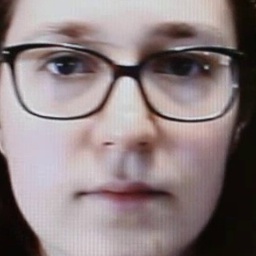}}\break
    \subfigure[]{\includegraphics[width=0.08\textwidth]{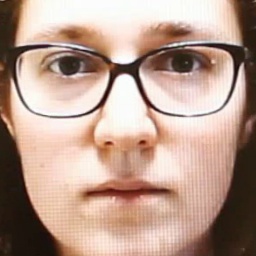}}
    \subfigure[]{\includegraphics[width=0.08\textwidth]{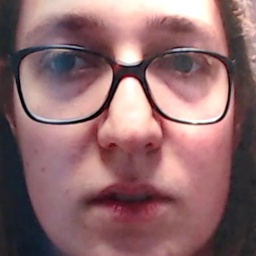}}
    \subfigure[]{\includegraphics[width=0.08\textwidth]{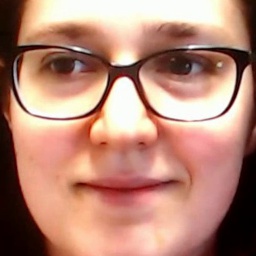}}
    \subfigure[]{\includegraphics[width=0.08\textwidth]{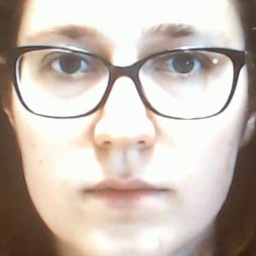}}
    \subfigure[]{\includegraphics[width=0.08\textwidth]{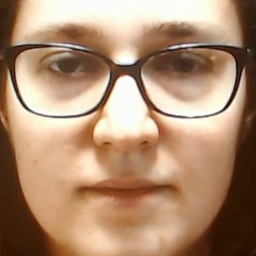}}

    \caption[]{Spoof representations of each type of presentation attack and image style present in the SiW database.}
    \label{fig:spoof-geradas}
\end{figure}

We used a CNN to perform style transfer based on just one reference image, following the implementation in \cite{git_style_engstrom2016faststyletransfer}. The CNN architecture replicates the VGG19 \cite{vgg19_Liu2015} architecture, with the parametrization, optimization and also the training method proposed in \cite{cnnopt1_gatys2015neural, cnnopt2_johnson2016perceptual, cnnopt3_ulyanov2016instance}, with perceptual loss and instance normalization. One subject of the database was randomly chosen and each one of its spoof distribution perceived was selected to be a reference image of spoof from the SiW database. Fig. \ref{fig:spoof-geradas} shows the 10 reference images of the spoof styles available in SiW.

\begin{figure*}[htbp!]
    \centerline{\includegraphics[width=14cm]{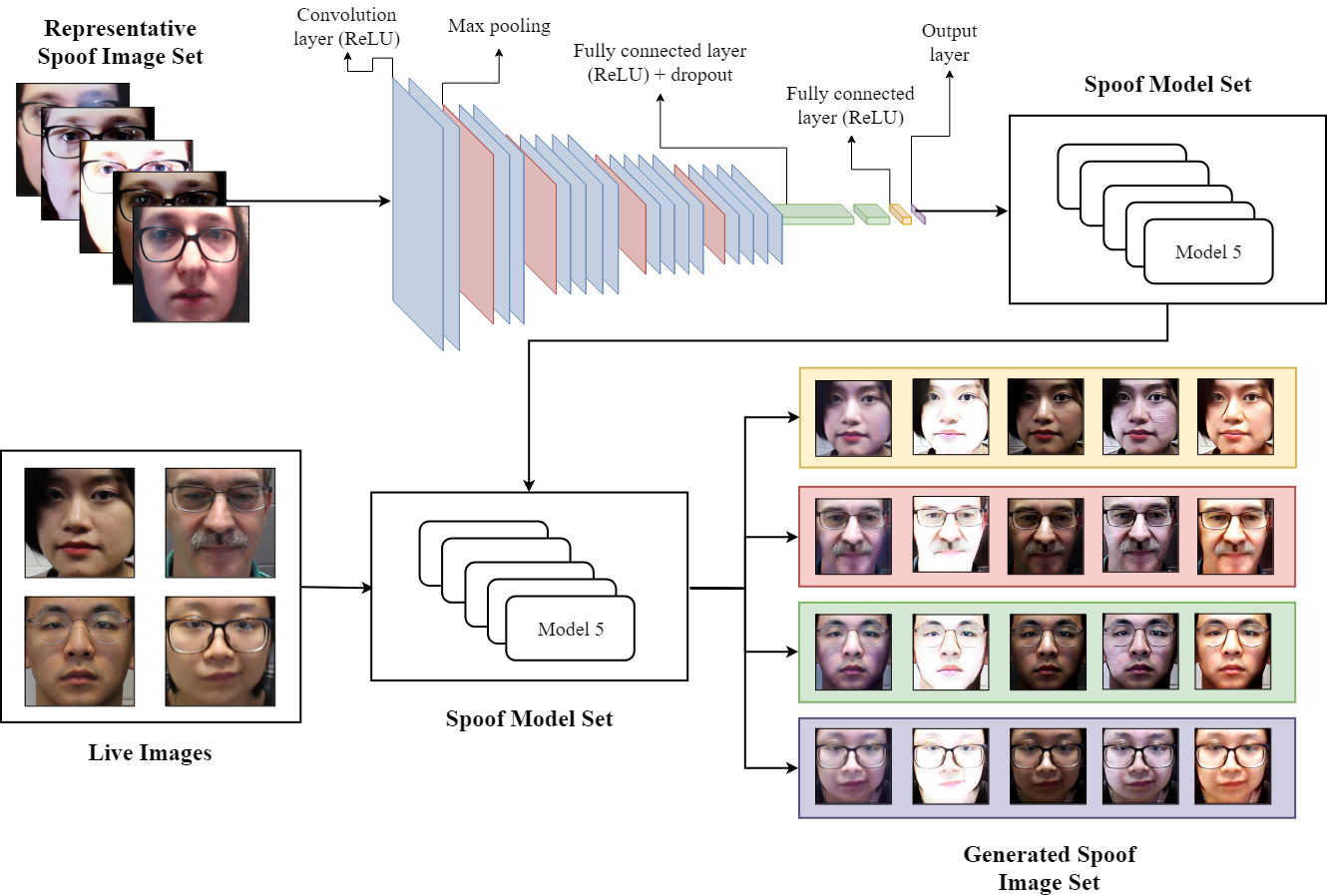}}
    \caption{Spoof image generation protocol pipeline. The VGG19 \cite{vgg19_Liu2015} architecture was used to obtain the style of the images. For each spoof representation, one image was used to generate a model capable of doing transfer style. For each subject, a set of spoof images was generated based on each of the spoof representations trained in the step previous explained.}
    \label{diag1}
\end{figure*}

Fig. \ref{diag1} presents the pipeline to generate spoof images using one image as a reference for each spoof style. First, a VGG19 \cite{vgg19_Liu2015} is used to extract information from the reference image, in order to obtain the stylization of the image. This step is performed for each spoof image representation, generating one model for each spoof style representation. Next, the spoof images are generated using each one of the spoof models. A holdout approach was used in the experiments and the database was split into 70\% of the live images for training and the remaining  30\% for test. Each training image is used as input to the style transfer CNN and it provides at the output 10 spoof samples, to maintain data balancing, just 10\% of the live training images were used during the process of spoof generation.

\subsubsection{Face Anti-Spoofing User-Centered Model (FAS-UCM)} 
\begin{figure}[h!]
    \centerline{\includegraphics[width=0.45\textwidth]{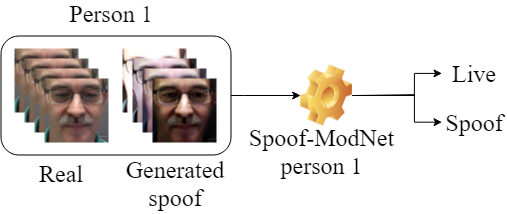}}
    \caption{Training process of a CNN for each subject. Two architectures were used, the proposed Spoof-ModNet and the pre-trained MobileNetV2.}
    \label{diag-dcnn-user}
\end{figure}

All generated spoof images and all live images from the training set are subsequently used to train the FAS-UCM to distinguish between live and spoof. Fig. \ref{diag-dcnn-user} illustrates the training process. We used two CNN architectures for the user-centered models: a MobileNetV2 CNN \cite{mobilenetv2_1801.04381}, pre-trained on the ImageNet \cite{imagenet_Deng2009} and the proposed Spoof-ModNet, as can be seen in Table \ref{table:shallownet}. While the Spoof-ModNet has a total number of parameters of 148k, the MobileNetV2 \cite{mobilenetv2_1801.04381} has between 1.7M to 6.9M parameters, which shows that the proposed model is substantial less complex. Also, in Table \ref{table:shallownet} is possible to see that the first convolutional layer takes as an input a 32x32 image, which significantly reduces the complexity of the neural network and speed-up the training and testing process. In contrast, the MobileNetV2 takes a 224x224 image as an input.


The MobileNetV2 architecture was fine-tuned with a learning rate of 0.01, batch size of 100 and 4000 steps. The proposed Spoof-ModNet was randomly initialized and trained with a learning rate of 0.0001, batch size of 8 and 50 epochs. The performance of both architectures was evaluated in our experiments.


\begin{table}[htbp]
\centering
\caption{Architecture of the proposed Spoof-ModNet used in the user-centered method. Two groups of conv (ReLU) and batch normalization followed by a max\_pooling and a dropout layer make up the structure of the network.}
\label{table:shallownet}
\renewcommand{\arraystretch}{1.3}
\begin{tabular}{|l|c|c|c|c|}
\hline
\textbf{layer} & \textbf{size-in}              & \textbf{size-out}             & \textbf{kernel}               & \textbf{params}                     \\ \hline
conv2d\_1      & 32$\times$32$\times$3                       & 32$\times$32$\times$16                      & 3$\times$3                           & 448                                 \\
activation\_1  & 32$\times$32$\times$16                      & 32$\times$32$\times$16                      &                               & 0                                   \\
batch\_norm\_1 & 32$\times$32$\times$16                      & 32$\times$32$\times$16                      &                               & 64                                  \\
conv2d\_2      & 32$\times$32$\times$16                      & 32$\times$32$\times$16                      & 3$\times$3                           & 2320                                \\
activation\_2  & 32$\times$32$\times$16                      & 32$\times$32$\times$16                      &                               & 0                                   \\
batch\_norm\_2 & 32$\times$32$\times$16                      & 32$\times$32$\times$16                      &                               & 64                                  \\
ma$\times$\_pool2d\_1 & 32$\times$32$\times$16                      & 16$\times$16$\times$16                      & 2$\times$2                           & 0                                   \\
dropout\_1     & 16$\times$16$\times$16                      & 16$\times$16$\times$16                      &                               & 0                                   \\
conv2d\_3      & 16$\times$16$\times$16                      & 16$\times$16$\times$32                      & 3$\times$3                           & 4640                                \\
activation\_3  & 16$\times$16$\times$32                      & 16$\times$16$\times$32                      &                               & 0                                   \\
batch\_norm\_3 & 16$\times$16$\times$32                      & 16$\times$16$\times$32                      &                               & 128                                 \\
conv2d\_4      & 16$\times$16$\times$32                      & 16$\times$16$\times$32                      & 3$\times$3                           & 9248                                \\
activation\_4  & 16$\times$16$\times$32                      & 16$\times$16$\times$32                      &                               & 0                                   \\
batch\_norm\_4 & 16$\times$16$\times$32                      & 16$\times$16$\times$32                      &                               & 128                                 \\
ma$\times$\_pool2d\_2 & 16$\times$16$\times$32                      & 8$\times$8$\times$32                        & 2$\times$2                           & 0                                   \\
dropout\_2     & 8$\times$8$\times$32                        & 8$\times$8$\times$32                        &                               & 0                                   \\
flatten\_1     & 8$\times$8$\times$32                        & 1$\times$1$\times$2048                      &                               & 0                                   \\
dense\_1       & 1$\times$1$\times$2048                      & 1$\times$1$\times$64                        &                               & 131136                              \\
activation\_5  & 1$\times$1$\times$64                        & 1$\times$1$\times$64                        &                               & 0                                   \\
batch\_norm\_5 & 1$\times$1$\times$64                        & 1$\times$1$\times$64                        &                               & 256                                 \\
dropout\_3     & 1$\times$1$\times$64                        & 1$\times$1$\times$64                        &                               & 0                                   \\
dense\_2       & 1$\times$1$\times$2                         & 1$\times$1$\times$2                         &                               & 130                                 \\
activation\_6  & 1$\times$1$\times$2                         & 1$\times$1$\times$2                         &                               & 0                                   \\ \hline

\textbf{Total} & \multicolumn{1}{l|}{\textbf{}} & \multicolumn{1}{l|}{\textbf{}} & \multicolumn{1}{l|}{\textbf{}} & \multicolumn{1}{l|}{\textbf{148562}} \\ \hline
\end{tabular}
\end{table}

In the next section we present the evaluation of the proposed method on the live and real spoof testing images -- the spoof images present in the original database -- for each subject.

\section{Experiments and Results}
\begin{figure}[htbp]
    \centerline{\includegraphics[width=0.45\textwidth]{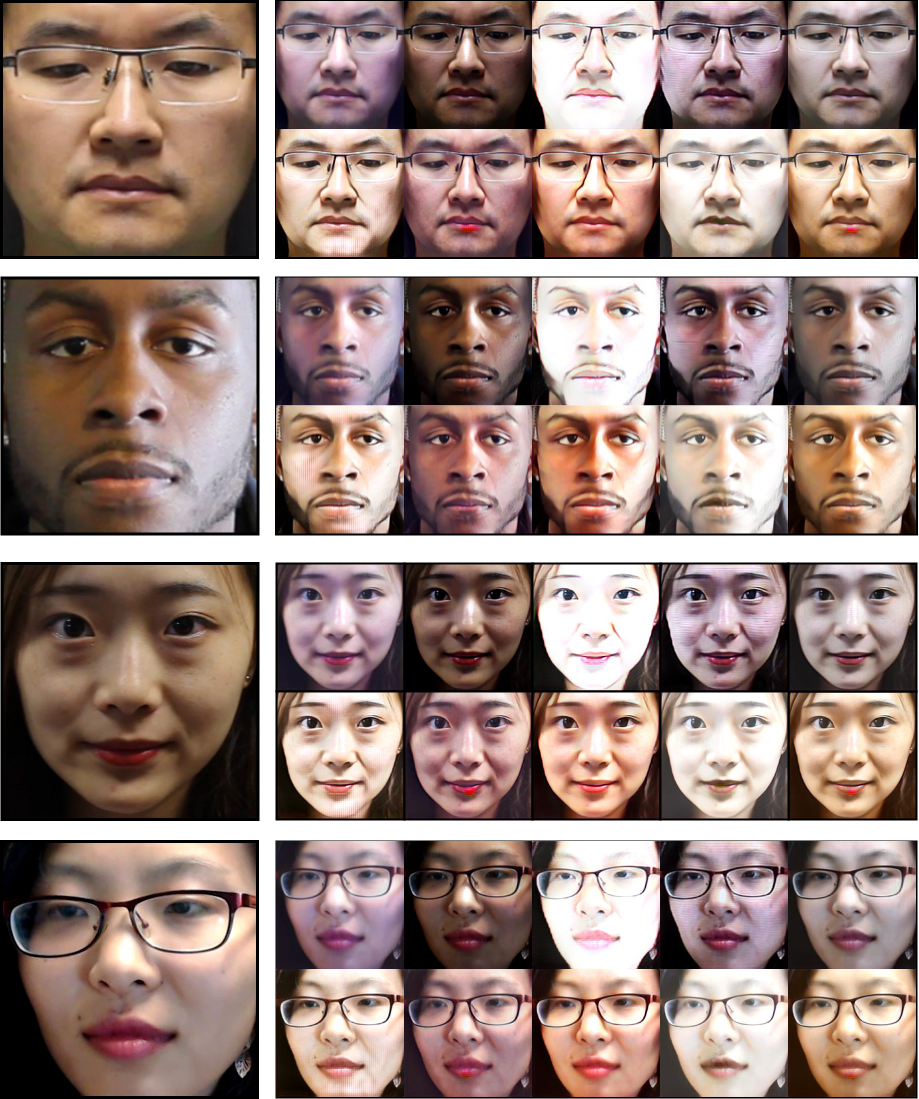}}
    \caption{Spoof images generated. On the left the live images and on the right the spoof images generated based on each of the spoof images representations of Fig. \ref{fig:spoof-geradas}.}
    \label{spoof-samples}
\end{figure}

The first important result obtained in this paper was the automatic generation of spoof images based on a single reference image. For each training image used as input on the style transfer CNN, 10 new spoof samples were generated. Examples of the generated spoof images can be seen in Fig \ref{spoof-samples}. The qualitative results show that the VGG19 network was able to capture and transfer the style of the spoof images. It is important to note that the network captured all the details present in the spoof images, for example very bright spots, warped lines, change in color distribution throughout the face image and change in the illumination. Another important aspect of the style transfer was the adaptation to subjects wearing glasses or not, given that all spoof image representations had a subject wearing glasses, which did not reflect in the resulting images. Also, it is possible to analyze the good adaptation of the CNN to the gender and race of the subject. 

The noise present in some of the generated images, usually close to the mouth and cheeks, can be explained as the CNN trying to transfer the very bright spots -- for example, specular reflections captured by the camera -- seen in the representative spoof images to an image that is not too similar from the reference image. Not far from that, noise can be perceived in some generated images with warped lines across the face, which is a result of the CNN trying to mimic the moir\'{e} pattern.

In order to evaluate the generalization of the CNNs trained with the generated spoof images, a testing protocol was applied using the test images. The classification test was performed for each subject, considering their respective previous trained model. Table \ref{tab:results1} shows the results over the two CNN architectures. The Spoof-ModNet has better performance over the metrics analyzed, with an average classification error rate (ACER) of 0.22, while the MobileNetV2 presented an ACER of 0.26. It is important to note that the Spoof-ModNet has significant less convolutional layers and the input images are also significantly smaller (32$\times$32$\times$3).

\begin{table}[h]
\centering
\renewcommand{\arraystretch}{1.3}
\caption{Results comparing the two architectures in the SiW database.}
\label{tab:results1}
\begin{tabular}{llllll}
\hline
\textbf{Architecture} & \textbf{Accuracy} & \textbf{FAR} & \textbf{FRR} & \textbf{F1 Score} & \textbf{ACER} \\ \hline
Spoof-ModNet           & 0.75            & 0.44         & 0.01         & 0.78          & 0.22        \\ 
MobileNetV2           & 0.71            & 0.53         & 0.00         & 0.75          & 0.26        \\ \hline
\multicolumn{6}{l}{\scriptsize FAR: False acceptance rate, FRR: False rejection rate} 
\end{tabular}
\end{table}

\begin{figure}[tbp]
    \centerline{\includegraphics[width=0.45\textwidth]{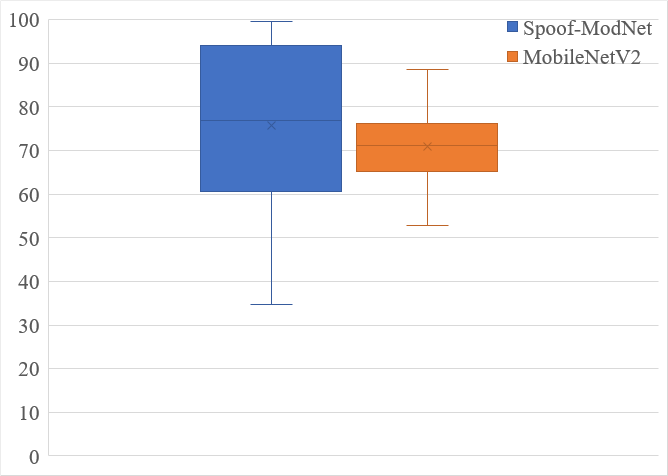}}
    \caption{Results in terms of accuracy for each subject grouped in a boxplot chart, considering both architectures.}
    \label{fig:resultados-por-pessoa}
\end{figure}

In Fig. \ref{fig:resultados-por-pessoa} it is possible to analyze the results considering the accuracy reported for each subject in the database. In total, there were 90 subjects in the SiW test database. The minimum accuracy reported with the proposed Spoof-ModNet was 34.69\%, and the maximum 99.49\%. Also, it is possible to see that more than 50\% of the reported accuracy per subject is above 70\% accuracy using the Spoof-ModNet, while 50\% of the data lies in the range of 60.49\% and 93.95\% accuracy. On the other hand, the MobileNetV2 presented a minimum of 52.90\% and a maximum of 88.53\% accuracy, with 50\% of the data between an accuracy of 65.07\% and 76.28\%. From the boxplot chart, it is also possible to observe that the MobileNetV2, despite having a worst average accuracy, had more consistent performance, varying less than the Spoof-ModNet architecture. 

For the performance evaluation and comparison with other published methods, besides the ACER metrics, we have also selected two other standardized ISO/IEC 30107-3 metrics \cite{ISO30107-3:2017}: attack presentation classification error rate (APCER) and normal presentation classification error rate (NPCER). We reported in Table \ref{tab:results2} an indirect comparison between our method and other methods \cite{siw_DBLP:journals/corr/abs-1803-11097, r40_DBLP:journals/corr/abs-1812-00408, r46_DBLP:journals/corr/abs-1811-05118, rOurs_DBLP:journals/corr/abs-1904-12490} with reported results in the SiW database. The results reported in \cite{siw_DBLP:journals/corr/abs-1803-11097, r40_DBLP:journals/corr/abs-1812-00408, r46_DBLP:journals/corr/abs-1811-05118, rOurs_DBLP:journals/corr/abs-1904-12490} followed the intra-database protocol 1 proposed by Liu et al. \cite{siw_DBLP:journals/corr/abs-1803-11097}, with a set of subjects in the training data and a distinct set of subjects in the testing data. However, given that our protocol relies on user-centered models, it is not possible to follow the same protocol and the results are not directly comparable, being used only as a baseline. Although our results yielded a worst perfomance, with error rates higher than the other approaches, it is significantly superior to the random classification and it is feasible for generating user's fraudulent images from real images and use them to train user-centered face liveness detection models. In addition, improvements in the models and methods for choosing the spoof representation images should improve the results.


\begin{table}[t]
\centering
\renewcommand{\arraystretch}{1.3}
\caption{Evaluation results of different methods in SiW.}
\label{tab:results2}
\begin{tabular}{lccc}
\hline
 & \textbf{APCER} & \textbf{NPCER} & \textbf{ACER} \\
\textbf{Method} & \textbf{(\%)} & \textbf{(\%)} & \textbf{(\%)} \\ \hline
FAS-BAS \cite{siw_DBLP:journals/corr/abs-1803-11097} & 3.58 & 3.58 & 3.58 \\
FAS-TD-SF \cite{r46_DBLP:journals/corr/abs-1811-05118} & 0.96 & 0.50 & 0.73 \\ 
FAS-TD-SF-CASIA-SURF \cite{r40_DBLP:journals/corr/abs-1812-00408} & 1.27 & 0.33 & 0.80 \\ 
Meta-FAS-DR \cite{rOurs_DBLP:journals/corr/abs-1904-12490} & 0.52 & 0.50 & 0.51 \\ 
\textbf{FAS-UCM (Ours)*} & \textbf{44.00} & \textbf{0.00} & \textbf{22.00} \\
\hline
\multicolumn{4}{l}{\scriptsize * Testing protocol differs from the others.}
\end{tabular}%
\end{table}

\section{Conclusions}

We have presented a method to use style transfer technique to generate spoof images. We used the VGG19 network, which was able to capture and transfer the style from spoof representations. The generate spoof images were used to train two different architectures for each person to perform liveness detection, the MobileNetV2, and the proposed Spoof-ModNet. The Spoof-ModNet network had better performance, with an ACER 0.22, while the MobileNetV2 presented an ACER of 0.26.

Further work will be done to explore other classifier architectures and even the combination of multiple classifiers. Further analysis will be done in other databases to evaluate the generalization of the proposed method. It will also be evaluated other methods for choosing the spoof representation images. Choosing the spoof representation based on the subject particularities may bring better results than using the same representation for the entire database.

\section*{Acknowledgment}

This work was supported by UNIVISION INFORMATICA LTDA and CNPq (National Council for Scientific and Technological Development) grant 306688/2018-2 and DAI CP n$^{\circ}$ 23/2018.

\balance



\end{document}